\documentclass[nonatbib]{elsarticle}

\makeatletter
\let\c@author\relax
\makeatother

\usepackage{amsmath, amssymb} 
\usepackage{amsthm} 
\usepackage{stmaryrd} 
\usepackage{bm}
\usepackage{empheq}

\usepackage{multirow,makecell}

\usepackage{siunitx}

\usepackage{hyperref}
\usepackage{cleveref}

\usepackage{caption}
\usepackage{subcaption}

\usepackage{graphicx}

\usepackage{tikz, tikz-3dplot}
\usepackage{pgfplots}
\usetikzlibrary{patterns}

\usepackage{biblatex}
\AtEveryBibitem{%
\ifentrytype{inproceedings}{%
    \clearname{editor}%
    \clearfield{eventdate}%
    \clearfield{eventyear}%
    \clearlist{publisher}%
    \clearlist{location}%
    \iffieldundef{doi}{}{\clearfield{url}}%
}{}
\ifentrytype{article}{%
    \clearfield{url}%
    \clearfield{eprint}%
}{}
\ifentrytype{incollection}{%
    \clearfield{url}%
    \clearfield{eprint}%
    \clearname{editor}%
    \clearfield{address}%
}{}
\ifentrytype{misc}{%
    \clearfield{url}%
    \clearfield{doi}%
}{}
}

\usepackage{color}
\usepackage{soul} 
\usepackage{todonotes} 

\SetSymbolFont{stmry}{bold}{U}{stmry}{b}{n}

\makeatletter
\if@todonotes@disabled

\else

\fi
\makeatother

\theoremstyle{definition}

\newcommand{\tr}[1]{\mathrm{tr}\left(#1\right)} 
\newcommand{\trp}{{\mathrm{T}}} 
\newcommand{\mandel}[3][]{\ifthenelse{\equal{#1}{}}%
    {\mathrm{M}_{#2}^{#3}}%
    {\mathrm{M}_{#1}^{#2}\left(#3\right)}} 
\newcommand{\diag}[1]{\mathrm{diag}\left(#1\right)} 

\newcommand{\iran}[2]{\left\llbracket#1, #2\right\rrbracket} 
\newcommand{\sym}[1]{\mathrm{S}^{#1}} 
\newcommand{\grp}[1][]{\ifthenelse{\equal{#1}{}}{\mathcal{G}}{\mathcal{G}_{#1}}} 
\newcommand{\og}[1]{\mathcal{O}\left(#1\right)} 
\newcommand{\tensor}[2]{\mathcal{T}_{#1}^{#2}} 

\newcommand{\eye}[1]{\bm{I}_{#1}} 

\date{}

\begin{document}

\title{Symmetry-enforcing neural networks with applications to constitutive modeling}

\author[uci]{K\'evin Garanger}
\ead{kevin.garanger@uci.edu}
\author[gatech]{Julie Kraus}
\ead{jkraus@gatech.edu}
\author[uci]{Julian J. Rimoli\corref{cor}}
\ead{jrimoli@uci.edu}

\cortext[cor]{Corresponding author}
\address[uci]{Department of Mechanical and Aerospace Engineering, University of California, Irvine, CA 92697, USA}
\address[gatech]{School of Aerospace Engineering, Georgia Institute of Technology, Atlanta, GA 30332, USA}

\begin{abstract}
The use of machine learning techniques to homogenize the effective behavior of arbitrary microstructures has been shown to be not only efficient but also accurate.
In a recent work, we demonstrated how to combine state-of-the-art micromechanical modeling and advanced machine learning techniques to homogenize complex microstructures exhibiting non-linear and history dependent behaviors~\cite{logarzo2021smart}.
The resulting homogenized model, termed smart constitutive law (SCL), enables the adoption of microstructurally informed constitutive laws into finite element solvers at a fraction of the computational cost required by traditional concurrent multiscale approaches.
In this work, the capabilities of SCLs are expanded via the introduction of a novel methodology that enforces material symmetries at the neuron level, applicable across various neural network architectures.
This approach utilizes tensor-based features in neural networks, facilitating the concise and accurate representation of symmetry-preserving operations, and is general enough to be extend to problems beyond constitutive modeling.
Details on the construction of these tensor-based neural networks and their application in learning constitutive laws are presented for both elastic and inelastic materials.
The superiority of this approach over traditional neural networks is demonstrated in scenarios with limited data and strong symmetries, through comprehensive testing on various materials, including isotropic neo-Hookean materials and tensegrity lattice metamaterials.
This work is concluded by a discussion on the potential of this methodology to discover symmetry bases in materials and by an outline of future research directions.
\end{abstract}

\maketitle

\section{Introduction}

The determination of the effective mechanical properties of materials is a vast topic that has historically been essentially based on experimental methods~\cite{dally2005experimental}, coupled with constitutive models specifically conceived for the type of material studied~\cite{ottosen2005mechanics}.
These constitutive models are fundamental to engineering-level analysis of the mechanics of materials, via their use in finite element methods~\cite{deborst2012nonlinear}.
Traditionally, they have been grouped into two categories --- phenomenological and micro-mechanical ---, although the distinction between the two is not always clearly defined.
The former type of models tends to rely on empirical considerations and observations, while the latter type is based on physical considerations of the material behavior.
What makes the distinction between the two types sometimes ambiguous is the fact that to be of practical use, even micro-mechanical models need to rely on some assumptions or simplifications.
Some examples of phenomenological models include~\cite{leclercq1996general,johnson1999response,deshpande2000isotropic,koo2010dynamic} while examples of micro-mechanical models are given in~\cite{hashin1963variational,hill1963properties,sun1991micromechanical,deshpande2008inelastic,angioni2011comparison}.

A different approach consists of using computational modeling of the material behavior at the microscale.
Various methods fall under this approach, such as computational homogenization, which is based on the principle of separation of scales
~\cite{guedes1990preprocessing,yvonnet2009numerically,leuschner2017reduced,,geers2017homogenization}, and concurrent multiscale methods, where a strong coupling between scales is considered~\cite{broughton1999concurrent,fish2000multiscale,zhang2011new,casadei2013geometric}.
A major drawback of these approaches is their computational cost, which tends to be prohibitively high for most three-dimensional problems~\cite{geers2017homogenization}, preventing their widespread use in engineering applications.

With the hope of benefiting from the efficiency of constitutive modeling and the accuracy of computational microscale modeling, data-driven methods based on machine learning have been proposed~\cite{amsallem2008interpolation,yvonnet2015computation,ibanez2016manifold,lu2019datadriven,wu2020recurrent,logarzo2021smart,khoei2023machine}.
With these methods, mathematical models capable of approximating complex relationships, such as artificial neural networks, are trained from data generated by computational simulations.
The use of very general and powerful models is justified by the need to capture intricate behaviors of materials that would be otherwise hard to efficiently model accurately from physical considerations only.
Conceptually, this approach differs significantly from the derivation of a constitutive model based on physics or empiricism that already captures some characteristics of the material response by nature.
This stark difference remains true even if some of these latter constitutive models are calibrated on data samples with methods inspired from machine learning, such as in~\cite{pal1996calibration}.

Artificial neural networks were proposed decades ago to model the mechanical response of materials such as concrete~\cite{ghaboussi1991knowledge}, sand~\cite{ellis1992neural}, or sandstone~\cite{millar1994investigation}, although these initial approach relied on a limited amount of experimental data to calibrate relatively small networks.
Subsequent early works revolving around neural networks for constitutive modelings include~\cite{furukawa1998implicit,lefik2003artificial,shen2004neural}.
More recently, thanks to advances in micro-structure simulation and the increasing availability of computational resources, larger and more powerful models have been used to learn constitutive laws from simulated data for elastic~\cite{yvonnet2015computation,lu2019datadriven,khoei2023machine} and inelastic materials~\cite{wu2020recurrent,logarzo2021smart,zhang2022learning}.
In these works, constitutive models are learned by training neural networks with standard deep learning methods on pairs of inputs and outputs, which consist respectively of kinematic variables (e.g., strain tensor or deformation gradient) and stress variables.

Despite these promising results, one of the disadvantages of neural networks for constitutive modeling is the fact that some of the properties that the material is known to satisfy, from mathematical or physical considerations, cannot in general be enforced.
These properties notably include hyperelasticity, frame indifference (or objectivity), material stability, invariance to material symmetries, and compliance with thermodynamics laws~\cite{masi2021thermodynamics}.
This drawback of neural networks is linked to their inherent ability to approximate arbitrary functions, and therefore can be seen as an inevitable consequence of this otherwise powerful and desirable feature.
To put in perspective this tension between the capability of a model to replicate intricate behavior and the intrinsic enforcement of known properties, micro-mechanical models can be thought as belonging to the other end of the spectrum, being derived from physical considerations but not always able to captures all aspects of a material's response.

In theory, a neural network large enough should be able to learn approximately the previously mentioned properties with sufficient data, but doing so may require an excessively large number of training samples and computational resources~\cite{cohen2016group}.
For material symmetry and frame indifference, the difficulty with obtaining a dataset large enough to implicitly represent these properties is partially alleviated by the use of data augmentation techniques.
With these techniques, new samples are generated from existing ones by applying a transformation that preserves the invariant properties of interest~\cite{shorten2019survey}.
However, data augmentation still does not solve the issue of computational cost and is not always applicable, as with the case of material symmetry if the basis of the material symmetry group is not known.
In addition, the approach consisting of learning some properties from a dataset may not be acceptable if these properties needs to be satisfied exactly, as such approximations may result in an inconsistent model that violates some laws of physics.
For example, the non-respect of frame indifference, itself a purely mathematical property, leads to a different work produced along a path depending on the frame of reference.

For these reasons, neural networks that are designed to satisfy these properties by construction have been proposed~\cite{ling2016machine,fernandez2020anisotropic,fernandez2021material,xu2021learning,asad2022mechanics,klein2022polyconvex}.
In addition to alleviate the already mentioned issues with data augmentation, neural networks that are designed to satisfy some of these properties tend to also perform better and require smaller models~\cite{ling2016machine,miller2020relevance}.
A common approach to guarantee hyperelasticity is to learn a strain energy function with a neural network potential~\cite{fernandez2020anisotropic,fernandez2021material,asad2022mechanics,klein2022polyconvex}, rather than a strain-stress relationship, similarly to what has been done in molecular chemistry~\cite{behler2015constructing}.
The frame indifference problem is generally solved by using an input representation that is independent of the frame of reference, such as the Cauchy-Green strain tensor~\cite{fernandez2020anisotropic,fernandez2021material} or Green-Lagrange strain tensor~\cite{asad2022mechanics}.
However, the use of these tensors as inputs is incompatible with the enforcement of material stability via input convex neural networks, which is a recent method that requires the deformation gradient as input~\cite{klein2022polyconvex}.
Finally, material symmetries have been enforced in previous works with one of two methods:
\begin{enumerate}
    \item By using invariants of the strain tensor as inputs~\cite{ling2016machine,klein2022polyconvex,linden2023neural},
    \item By using a process known as group symmetrization, where all the transformed versions of an input with respect to a symmetry group are passed through a neural network and averaged~\cite{fernandez2020anisotropic,fernandez2021material,klein2022polyconvex}.
\end{enumerate}
One of the advantages of the first method compared to the second one is that the basis of the materials symmetry group does not need to be known.
It can also be applied to infinite symmetry group, such as transverse isotropy~\cite{fernandez2021material}.
The second method can be adapted to infinite groups by considering only a finite subgroup of the materials symmetries, but this results in the material symmetries being only approximately satisfied by the trained constitutive law.
In addition, the larger the symmetry group, the more neural network outputs need to be computed with the second method, making it computationally more expensive than the first one for large groups.
On the other hand, the first method requires identifying a set of invariants that is suited to the problem, which can be challenging, given that many different invariants exist~\cite{steigmann2003isotropic,schroder2005variational,balzani2006polyconvex,ebbing2010construction} and not all of them are appropriate for numerical computations~\cite{ta2014constructive}.
What's more, there also seems to be cases where learning constitutive laws from invariants fails, unlike when learning from the full strain tensor~\cite{klein2022polyconvex}.

In this work, a new method to enforce materials symmetry is proposed, in which the materials symmetries are exactly enforced at the neuron level, independently of the neural network architecture.
This method relies on linear operations and activations that are equivariant to the action of the material symmetries.
With traditional neural networks, features describe each individual scalar entry of a network's inputs and outputs, as well as the hidden layers' outputs.
With the proposed approach, this concept is extended to tensors, by considering features represented by tensors.
While the inputs and outputs of the networks introduced here are each individual tensors (a strain tensor and a stress tensor), the hidden layers' outputs are represented as arrays of tensors.
This formalism does not change in essence the nature of the linear operations occurring between layers, but it allows to conveniently write symmetry-preserving operations in a concise and straightforward way, while reusing known results from linear elasticity applied to materials with symmetries.
However, the proposed activation functions, since they are applied feature-wise, are not equivalent to activation layers of standard neural networks, which are applied to each scalar individually.
The methodology to construct neural networks based on tensor features are detailed in~\cref{sec:methodology}.
The resulting neural networks are named \emph{tensor feature equivariant neural networks} (TFENN).
Although the examples presented in this work come from continuum mechanics, the proposed methodology is general and can be applied to any problem where the inputs and outputs can be represented as collections of symmetric tensors.

Both feed-forward and recurrent TFENNs are presented, allowing the learning of constitutive models from elastic and inelastic materials.
To apply the proposed method, a basis of the symmetry must be known, which at first may seem like a drawback in comparison to invariants-based methods.
However, this requirement can be turned into an advantage by considering a symmetry basis as a learnable parameter of the neural network, therefore allowing its discovery through training.
Only rate-independent problems are considered in this work, but the methodology is easily extended to viscoplasticity, for which the formulation of material symmetries is essentially the same.
In~\cref{sec:results}, the proposed methodology is applied to the learning of constitutive laws for several materials:
a three-dimensional isotropic neo-Hookean material, the unit cell of a two-dimensional tensegrity lattice metamaterial, and a two-dimensional plastic microstructure.
Results indicate that TFENNs are capable of learning complex constitutive models despite the apparently restrictive constraints placed on them.
In addition, they tend to outperform standard neural networks with similar numbers of parameters and same numbers of layers, with the most significant improvements being for scenarios with scarce data and strong symmetries.
The ability of standard neural networks to learn a dataset's symmetries is also investigated via numerical comparisons with TFENNs.
An example of symmetry basis discovery is also presented, based on the tensegrity lattice metamaterial dataset.
In addition, some of challenges and limitations of the proposed methodology are discussed.
Finally, in~\cref{sec:conclusion}, main conclusions of the study are provided, as well as potential future research directions.

\section*{Notations}
The following notations are used throughout the paper:
\begin{itemize}
    \item $\mathbb{R}$: the set of real numbers.
    \item $\iran{i}{j}$: the set of integers $\{i, i+1, \ldots, j\}$.
    \item $\sym{n}$: the set of symmetric matrices of size $n \times n$.
    \item $\og{n}$: the set of orthogonal matrices of size $n \times n$.
    \item $\tensor{n}{d}$: the set of real tensors of order $d$ and dimensions all equal to $n$.
    \item $\diag{v}$: the diagonal matrix with diagonal entries given by the vector $v$.
    \item $\tr{A}$: the trace of the matrix $A$.
    \item $\eye{n}$: the identity matrix of size $n$.
    \item $\bm{A} : \bm{B}$: the double contraction between two tensors of order at least 2 $\bm{A}$ and $\bm{B}$.
\end{itemize}

\section{Neural networks with tensor features}\label{sec:methodology}

\subsection{Equivariant feedforward neural networks}\label{sec:methodology:ff_nn}

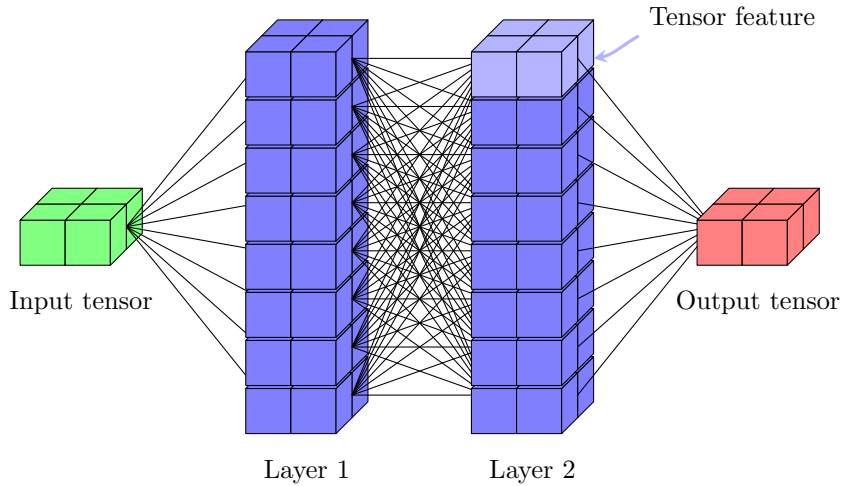
\begin{figure}[h]
  \centering
  \begin{tikzpicture}[x={(-0.35355cm,-0.35355cm)},y={(1cm,0cm)}, z={(0cm,1cm)}]

    \def\l{3} 
    \def\h{0.6} 
    \def\d{2} 
    \def\ln{8, 8} 
    \def\hl{{2, 8}} 
    \def\co{0.0} 
    \def\fs{0.04} 
    \def\bbo{0.4} 

      \tikzset{%
        pics/cube/.style n args={1}{%
          code={%
            \begin{scope}[fill,line join=round,line cap=round]
              \draw[fill] ({+0.5*#1}, {-0.5*#1}, {-0.5*#1}) coordinate (front-left-bottom) -- 
                ({+0.5*#1}, {+0.5*#1}, {-0.5*#1}) coordinate (front-right-bottom) -- 
                ({+0.5*#1}, {+0.5*#1}, {+0.5*#1}) coordinate (front-right-top) -- 
                ({+0.5*#1}, {-0.5*#1}, {+0.5*#1}) coordinate (front-left-top) -- cycle;
              \draw[fill] (front-right-top) --
                (front-right-bottom) --
                ({-0.5*#1}, {+0.5*#1}, {-0.5*#1}) coordinate (back-right-bottom) --
                ({-0.5*#1}, {+0.5*#1}, {+0.5*#1}) coordinate (back-right-top) -- cycle;
              \draw[fill] (back-right-top) --
                (front-right-top) --
                (front-left-top) --
                ({-0.5*#1}, {-0.5*#1}, {+0.5*#1}) coordinate (back-left-top) -- cycle;
            \end{scope}
          }
        }
      }

      \tikzset{%
        pics/tensor_feature/.style n args={2}{%
          code={%
            \foreach \y in {1, ..., #1} {%
              \foreach \x in {1, ..., #1} {%
                \pic at ({(\x-0.5*(#1+1))*#2}, {(\y-0.5*(#1+1))*#2}, 0) {cube={#2}};
              };
            };
          }
        }
      }

      \begin{scope}[local bounding box=input]
        \pic[black,fill=white!50!green] at (0, 0 0) {tensor_feature={\d}{\h}};
      \end{scope}
      \node at ($(input.south) - (0,0,0.5)$) {Input tensor};

      \def\prevn{1}

      \foreach[expand list] \n [count=\yi] in {\ln}
      {%
        \foreach \z in {1, ..., \n}
          {%
            \foreach \prevz in {1, ..., \prevn}
              {
                \draw[black] (0, {(\yi-1)*\l + \d*\h * (0.5+\co)}, {(\prevz - 0.5*(\prevn+1)) * (\h+\fs)}) --
                  (0, {\yi*\l - \d*\h * (0.5+\co)}, {(\z - 0.5*(\n+1)) * (\h+\fs)});
              };
          };
        \begin{scope}[local bounding box=layer\yi]
          \foreach \z in {1, ..., \n}
            {%
              \pgfmathparse{\yi == \hl[0] && \z == \hl[1] ? 1 : 0}
              \ifnum\pgfmathresult>0\relax%
                \begin{scope}[local bounding box=feature_hl]
                  \pic[black,fill=white!70!blue] at (0, {\yi*\l}, {(\z - 0.5*(\n+1))*(\h+\fs)}) {tensor_feature={\d}{\h}};
                \end{scope}
              \else%
                \pic[black,fill=white!50!blue] at (0, {\yi*\l}, {(\z - 0.5*(\n+1))*(\h+\fs)}) {tensor_feature={\d}{\h}};
              \fi
            };
            \xdef\prevn{\n}
        \end{scope}
        \node at ($(layer\yi.south) - (0,0,0.5)$) {Layer \yi};
      };

    \draw[-stealth,draw=white!70!blue,very thick] ($(feature_hl.east) + (0, \l * 0.2, \l * 0.1)$) node[above right] {Tensor feature} to[in=10,out=210] (feature_hl.east);

    \pgfmathparse{dim{{\ln}}+1}

    \foreach \prevz in {1, ..., \prevn}
      {%
        \draw[black] (0, {(\pgfmathresult-1)*\l + \d*\h * (0.5+\co)}, {(\prevz - 0.5*(\prevn+1))*(\h+\fs)}) --
          (0, {\pgfmathresult*\l - \d*\h * (0.5+\co)}, {0});
      };

      \begin{scope}[local bounding box=output]
        \pic[black,fill=white!50!red] at (0, {\pgfmathresult*\l}, 0) {tensor_feature={\d}{\h}};
      \end{scope}
      \node at ($(output.south) - (0,0,0.5)$) {Output tensor};

  \end{tikzpicture}
    \caption{Representation of a TFENN with two hidden layers of eight two-dimensional tensor features each.
        Each connection between two features is associated to a weight tensor and each feature is associated with a bias tensor.
        The weights of the connections leading to a given feature, the bias tensor of this feature, and the activation at the corresponding layer fully characterize a neuron operation.
    }%
    \label{fig:tensor_feature_layer}
\end{figure}

TFENNs are meant to learn a constitutive relationship $f$ between a pair of second-order tensors of same size, the first one being a kinematic input and the second one a stress output.
In particular, since the primary focus of this work is the enforcement of material symmetries, the input and output tensors are assumed to be symmetric tensors $\bm{M} \in \sym{d}$ that, for any matrix $\bm{Q} \in \og{d}$, are transformed via the group action: $\bm{M} \longmapsto \bm{Q}^\trp \bm{M} \bm{Q}$, where $d$ is the appropriate dimension.
The Cauchy-Green strain tensor $\bm{C}$ and the second Kirchoff-Piola stress tensor $\bm{S}$, which are work conjugates up to a multiplicative constant, are an example of a pair of such tensors.
For a history-independent material, using these tensors as input and output to a neural network allows to formulate the material symmetries of the constitutive relationship as
\begin{equation}\label{eq:constitutive_model_symmetry}
    \forall \bm{Q} \in \grp, \forall \bm{C} \in \sym{d}, \quad f(\bm{Q}^\trp \bm{C} \bm{Q}) = \bm{Q}^\trp f(\bm{C}) \bm{Q},
\end{equation}
where $\grp$ is the group of material symmetries, a subgroup of $\og{d}$.
If $f$ satisfies \cref{eq:constitutive_model_symmetry}, it is said to be $\grp$-equivariant, which is the property that is sought of TFENNs.
To construct such neural networks, symmetric tensors are used to represent intermediate features resulting from internal operations of the network (see \cref{fig:tensor_feature_layer}), and said operations are designed to be $\grp$-equivariant themselves.

\subsection{Neuron operations}\label{sec:methodology:ff_nn:neuron_operations}

At each layer of the proposed neural networks, a set of neuron operations is applied to a set of symmetric tensor features to produce a new set of symmetric tensor features.
Assuming that a layer has a number of input (respectively output) features $n_{\mathrm{in}}$ (respectively $n_{\mathrm{out}}$), and that the input (respectively output) features are denoted by $\bm{x}_j$ (respectively $\bm{h}_i$), the neuron operations is expressed as
\begin{equation}\label{eq:neuron_operations}
    \forall i \in \iran{1}{n_{\mathrm{out}}}, \quad \bm{h}_i = \phi\left(\sum_{j=1}^{n_{\mathrm{in}}} \bm{W}_{ij} : \bm{x}_j + \bm{b}_i\right).
\end{equation}
In \cref{eq:neuron_operations}, $\bm{W}_{ij} \in \tensor{d}{4}$ is a \emph{weight} tensor with minor symmetries, $\bm{b}_i \in \sym{d}$ is a \emph{bias} tensor, and $\phi : \sym{d} \longmapsto \sym{d}$ is a \emph{tensorial activation function}.
To ensure that the neuron operations are equivariant, each weight $\bm{W}$, each bias $\bm{b}$, and the activation $\phi$ are chosen such that
\begin{subequations}
    \begin{empheq}[left={\forall\bm{Q} \in\grp, \forall\bm{x} \in\sym{d}, \quad \empheqlbrace}]{align}
        \bm{W}:\left(\bm{Q}^\trp\bm{x} \bm{Q}\right) &= \bm{Q}^\trp\left(\bm{W}:\bm{x} \right) \bm{Q},\label{eq:neuron_invariance:weight} \\
        \bm{b} &= \bm{Q}^\trp\bm{b} \bm{Q},\label{eq:neuron_invariance:bias} \\
            \phi(\bm{Q}^\trp\bm{x} \bm{Q}) &= \bm{Q}^\trp\phi(\bm{x}) \bm{Q}.\label{eq:neuron_invariance:activation}
    \end{empheq}
\end{subequations}

Clearly, \cref{eq:neuron_invariance:weight,eq:neuron_invariance:bias,eq:neuron_invariance:activation} are sufficient conditions for the equivariance of the neural network resulting from the composition of multiple layers.

\subsubsection{Weights}\label{sec:methodology:ff_nn:weights}

The form that $\bm{W}$ must take to satisfy \cref{eq:neuron_invariance:weight} given $\grp$ is a well-studied problem arising from the characterization of the elasticity tensor of linear elastic materials with symmetries.
It was shown that the number of different equivalence classes for material group symmetries in two dimensions is \num{4} and \num{6} for hyperelastic and elastic materials respectively~\cite{he1996symmetries}, and \num{8}~\cite{forte1996symmetry} and \num{12}~\cite{yongzhong1991completeness} in three dimensions.
The tensor forms corresponding to each of these classes in the natural basis of symmetry are given in multiple references, for example in~\cite{auffray2016handbook} for the 2D case and~\cite{abramian2020recovering} for the 3D case.

\subsubsection{Biases}\label{sec:methodology:ff_nn:biases}

If $\grp$ is the trivial group, i.e.\ triclinic symmetry or no symmetry is assumed, any symmetric tensor $\bm{b}$ can be used to satisfy \cref{eq:neuron_invariance:bias}.
Otherwise, the form of $\bm{b}$ is more constrained and but can be determined for each symmetry group easily by solving a set of linear systems.
For example, with isotropic or cubic symmetry, $\bm{b}$ must be a multiple of the identity, while with orthotropic symmetry and $d=2$, it must be diagonal.

\subsubsection{Activations}\label{sec:methodology:ff_nn:activations}

Finally, to satisfy \cref{eq:neuron_invariance:activation}, the approach proposed in this work is to define $\phi$ such that it applies a \emph{scalar} activation function $\sigma$ to the eigenvalues $\bm{\lambda}$ of its argument.
This is done by defining the mapping $\Phi$ from scalar activations to tensorial activations such that $\forall \bm{U} \in \og{d}, \forall \bm{\lambda} \in \mathbb{R}^d$,
\begin{equation}\label{eq:ev_tensorial_activation}
    \Phi(\sigma)\left(
        \bm{U}\diag{\bm{\lambda}} \bm{U}^\trp
    \right) 
    =
    \bm{U}
    \diag{\begin{bmatrix}
            \sigma(\bm{\lambda}_1) & \cdots & \sigma(\bm{\lambda}_d)
        \end{bmatrix}}
    \bm{U}^\trp.
\end{equation}
Given \cref{eq:ev_tensorial_activation}, it is obvious that $\Phi$ is well-defined (i.e.\ does not depend on the choice of $\bm{U}$) and that $\phi := \Phi\left(\sigma\right)$ satisfies \cref{eq:neuron_invariance:activation}.

\subsection{Equivariant recurrent neural networks}\label{sec:methodology:rnn}

The concept of tensor features can be extended to recurrent neural networks (RNNs) to model history-dependent constitutive models.

\subsubsection{Standard neuron operations}\label{sec:methodology:rnn:neuron_operations}

The most popular RNN cells, such as the LSTM~\cite{hochreiter1997long} and the GRU~\cite{cho2014properties}, are based on standard neuron operations.
Some of these neurons output intermediate features and some are \emph{gates} that control the flow of information through the network.
These gates use logistic activation functions and their output is multiplied by the output of other neurons.
For example, a possible layer operation of a GRU with scalar features is
\begin{align}
    \bm{r} &= \sigma\left(\bm{w}^{\mathrm{ir}} \bm{x} + \bm{w}^{\mathrm{hr}} \bm{h}' + \bm{b}^{\mathrm{r}}\right), \label{eq:gru:r} \\
    \bm{z} &= \sigma\left(\bm{w}^{\mathrm{iz}} \bm{x} + \bm{w}^{\mathrm{hz}} \bm{h}' + \bm{b}^{\mathrm{z}}\right), \label{eq:gru:z} \\
    \hat{\bm{h}} &= \psi\left(\bm{w}^{\mathrm{ih}} \bm{x} + \bm{b}^{\mathrm{ih}} + \bm{r} \odot \left(\bm{w}^{\mathrm{hh}}  \bm{h}' + \bm{b}^{\mathrm{hh}}\right)\right), \label{eq:gru:n} \\
    \bm{h} &= \left(1 - \bm{z}\right) \odot \hat{\bm{h}} + \bm{z} \odot \bm{h}' \label{eq:gru:h},
\end{align}
where $\bm{x}$ is the vector of input scalar features, $\bm{h}'$ is the vector of output scalar features at the previous time step, $\sigma$ and $\psi$ are activation functions applied element-wise (respectively a logistic function and a hyperbolic tangent), $\bm{r}$ and $\bm{z}$ are gate output, $\hat{\bm{h}}$ is a candidate output, and the various $\bm{w}$ and $\bm{b}$ are respectively weight matrices and bias vectors.
In \cref{eq:gru:n,eq:gru:h}, $\odot$ denotes the element-wise product of vectors.

\subsubsection{Tensorial neuron operations}\label{sec:methodology:rnn:tensorial_neuron_operations}

To extend this architecture to tensor features, the following modifications are proposed:
\begin{itemize}
    \item The input and output of the layer are replaced with arrays of tensor features, similarly to the approach proposed for feedforward neural networks.
    \item The output of gates are kept as vectors of scalars to remain faithful to the original purpose of gates which is to keep or forget specific features.
    \item The activation of \cref{eq:gru:n} is replaced with a tensorial activation function.
    \item The operation $\odot$ is replaced with an element-wise product between vectors of scalars and arrays of tensors.
\end{itemize}
These modifications lead to the following operations, written at the neuron level:
\begin{align}
    \bm{r}_i = \sigma\left(\sum_{j=1}^{n_\mathrm{in}} \tr{\bm{w}^{\mathrm{ir}}_{ij} \bm{x}_j} + \sum_{j=1}^{n_\mathrm{out}} \tr{\bm{w}^{\mathrm{hr}}_{ij} \bm{h}_j'} + \bm{b}^\mathrm{r}_i\right), \label{eq:gru_tensorial:r} \\
    \bm{z}_i = \sigma\left(\sum_{j=1}^{n_\mathrm{in}} \tr{\bm{w}^{\mathrm{iz}}_{ij} \bm{x}_j} + \sum_{j=1}^{n_\mathrm{out}} \tr{\bm{w}^{\mathrm{hz}}_{ij} \bm{h}_j'} + \bm{b}^\mathrm{z}_i\right), \label{eq:gru_tensorial:z} \\
    \hat{\bm{h}}_i = \Phi(\psi)\left(\sum_{j=1}^{n_\mathrm{in}} \bm{W}^{\mathrm{ih}}_{ij} : \bm{x}_j + \bm{b}^\mathrm{ih}_i + \bm{r}_i \left(\sum_{j=1}^{n_\mathrm{out}} \bm{W}^{\mathrm{hh}}_{ij} : \bm{h}_j' + \bm{b}^\mathrm{hh}_i\right) \right), \label{eq:gru_tensorial:n} \\
    \bm{h}_i = \left(\bm{1} - \bm{z}_i\right) \hat{\bm{h}}_i + \bm{z}_i \bm{h}_i', \label{eq:gru_tensorial:h}
\end{align}
where the various $\bm{W} \in \tensor{d}{4}$ are tensor weights, the $\bm{b} \in \sym{d}$ are bias tensors, and the $\bm{w} \in \sym{d}$ are matrix weights.
The linear operations in gates are from $\sym{d}$ to $\mathbb{R}$, which explains the use of the trace operator, since linear forms on $\sym{d}$ are isomorphic to $\sym{d}$ via the isomorphism $\bm{s} \in \sym{d} \longmapsto \tr{\bm{s}\ \cdot}$.

The equivariance of the neuron operation defined by \cref{eq:gru_tensorial:r,eq:gru_tensorial:z,eq:gru_tensorial:n,eq:gru_tensorial:h} can be ensured by choosing the weights and biases such that the operation in \cref{eq:gru_tensorial:n} is equivariant and the operations in \cref{eq:gru_tensorial:r,eq:gru_tensorial:z} are \emph{invariant}.
The invariance of these last operations means that a rotation of $\bm{x}$ and $\bm{h}'$ by the same orthogonal matrix results in unchanged output $\bm{r}$ and $\bm{z}$.
All these conditions are satisfied by choosing the tensor weights $\bm{W}$ as in \cref{eq:neuron_invariance:weight} and the bias tensors $\bm{b}$ and matrix weights $\bm{w}$ as in \cref{eq:neuron_invariance:bias}.
The fact that the invariance of the operations in \cref{eq:gru_tensorial:r,eq:gru_tensorial:z} requires this choice of weights is a consequence of the fact that $\forall \bm{Q} \in \og{d}, \tr{\bm{w} \bm{Q}^\trp \bm{x} \bm{Q}} = \tr{\bm{Q} \bm{w} \bm{Q}^\trp \bm{x}}$ and that $\forall \bm{y} \in \sym{d}, (\tr{\bm{x}\bm{y}} = 0\ \forall\bm{x} \in \sym{d} \implies \bm{y} = 0)$.

These conditions that the various $\bm{W}$, $\bm{w}$, $\bm{b}$ need to satisfy thus allows their full characterization as performed in \cref{sec:methodology:ff_nn:weights,sec:methodology:ff_nn:biases}.

\subsubsection{Other neural network architectures}\label{sec:methodology:rnn:other_neural_network_architectures}

Although the presented approach is illustrated with a GRU, it can be applied without difficulty to other recurrent neural network architectures.
This includes models with hidden state variables like LSTM by simply using tensor features for these variables.

\section{Results}\label{sec:results}

The results of applying TFENNs to multiple materials is presented in this section.
These materials include a neo-Hookean 3D hyperelastic material, a 2D unit of an elastic tensegrity metamaterial, and a 2D representative volume element of an elasto-plastic microstructure.
Standard neural network architectures with the same number of layers and a similar or higher number of parameters are also trained on these problems for comparison.

\subsection{Implementation details}\label{sec:results:implementation}

\subsubsection{Neural network implementation}\label{sec:results:implementation:nn}

All neural networks and training scripts are implemented in Python with the automatic differentiation package Jax~\cite{jax2018github} and other packages based on it~\cite{deepmind2020jax,flax2020github}.
An important detail is the fact that the tensorial activations as defined in \cref{eq:ev_tensorial_activation} are implemented via an eigenvalue decomposition of the input, which is only differentiable when eigenvalues are simple.
This initially led to numerical issues when attempting to compute the gradient of TFENNs.
However, using double precision floating point numbers and ensuring that no weights of the neural network are zero-initialized solved these issues in practice for the presented results.
A better option for future work would be to express \cref{eq:ev_tensorial_activation} (which itself is differentiable everywhere) as a composition of differentiable matrix functions.
For activations based on polynomial, exponential, or logarithm functions, this could be done by using the matrix versions of these functions.
However, not all of them are currently implemented in Jax.

\subsubsection{Random deformations generation}\label{sec:results:implementation:strain}

Random deformation gradients $\bm{F}$ are generated from positive eigenvalues sampled with a uniform distribution (hence defining a diagonal matrix $\bm{D}$) and from random rotation matrices $\bm{R}$, both combined by setting $\bm{F} = \bm{R} \bm{D}$.
In two dimensions, rotation matrices are generated by sampling random angles from a uniform distribution between $0$ and $2\pi$.
In three dimensions, they are constructed from random rotations axes sampled from a uniform distribution on the unit sphere and random rotation angles sampled as in two dimensions.

\subsubsection{Data scaling}\label{sec:results:implementation:scaling}

An important step in the training of neural networks is the scaling of the input and output data.
Here, three different scaling schemes are used:
\begin{enumerate}
  \item \textbf{Scalar component-wise scaling}: each component of the input and output tensors is scaled by its mean and standard deviation over the training set.
    This method is used to scale the input-output pairs provided during training to the standard neural networks, but is not suitable for the TFENNs, as it would break the possible symmetries of the data.
  \item \textbf{Tensor symmetry-preserving scaling}: input and output tensors are shifted and scaled by adding and multiplying them with multiples of the identity matrix, such that the mean and standard deviation of the diagonal coefficients of the scaled tensors are respectively zero and one. This scaling ensures that  material symmetries are preserved and is therefore used to scale the data provided to TFENNs.
  \item \textbf{Global scaling}: identical to the scalar component-wise scaling, except that the standard deviation is computed over all components of the tensors.
    This method is used for the computation of the mean squared error loss during training and validation to provide a same objective in all cases and a comparable outcome.
    In other words, after a neural network predicts a scaled output tensor according to the first or second scheme, this tensor is rescaled according to the third scheme before being compared to the target tensor, itself also scaled according to the third scheme.
    Given that the loss function involves a subtraction between the predicted and target tensors, the shifting of the tensors has no impact on the loss and only the scaling is relevant.
    The scaling involves a standard deviation computed over all components of the tensors to ensure that the components of the tensors are represented in the loss according to their original magnitude.
    All values of the validation loss reported in this section are computed using the third scheme.
\end{enumerate}

\subsubsection{Loss function}\label{sec:results:implementation:loss}

The loss function used to train all neural networks is the mean squared error between the predictions and targets.
That is, for a set of $N$ pairs of inputs and outputs $\{(\bm{X}_i, \bm{Y}_i)\}_{i=1}^N$, which are either tensors or sequences of tensors grouped in a single array, the loss function is defined as
\begin{equation}
  \mathcal{L} = \frac{1}{N} \sum_{i=1}^N \left\| \bm{Y}_i - \bm{f}(\bm{X}_i) \right\|_F^2,
\end{equation}
where $\left\| \cdot \right\|_F$ denotes the Frobenius norm.

\subsection{Neo-Hookean material}\label{sec:results:neo_hookean}

The first material considered is an isotropic neo-Hookean hyperelastic material.
The model used for this material is derived from the strain energy
\begin{equation}\label{eq:neo_hookean:strain_energy}
    W = \frac{\lambda}{2} {\left(\log{\det{\bm{F}}}\right)}^2
        - \mu \log{\det{\bm{F}}}
        + \frac{\mu}{2} \big(\tr{\bm{F}^\trp \bm{F}} - 3\big),
\end{equation}
where $\lambda$ and $\mu$ are model parameters.
\Cref{eq:neo_hookean:strain_energy} results in the following constitutive model,
expressed with respect to the Cauchy-Green and second Piola-Kirchoff tensors
$\bm{C}$ and $\bm{S}$:
\begin{equation}\label{eq:neo_hookean:constitutive_model}
    \bm{S} = \left(
            \frac{1}{2} \lambda \log{\det{\bm{C}}} - \mu 
        \right) \bm{C}^{-1}
        + \mu \eye{3}.
\end{equation}

\begin{figure}[h]
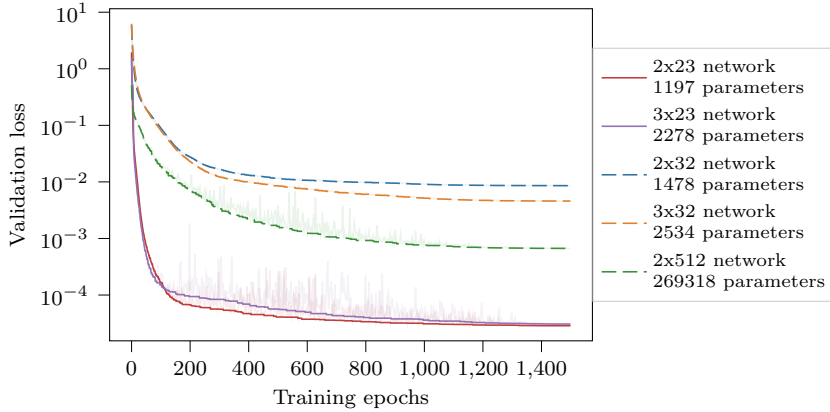

  \centering

  \caption{%
      \emph{Validation loss progress of neural networks trained on a neo-Hookean dataset with 20000 training samples.}
      The minimum validation loss up to the current epoch for tensor feature-based and standard networks is shown by solid and dashed lines, respectively.
      The current validation loss is represented by the dimmed lines.
  }%
  \label{fig:neohookean_training}
\end{figure}

\begin{figure}[h]
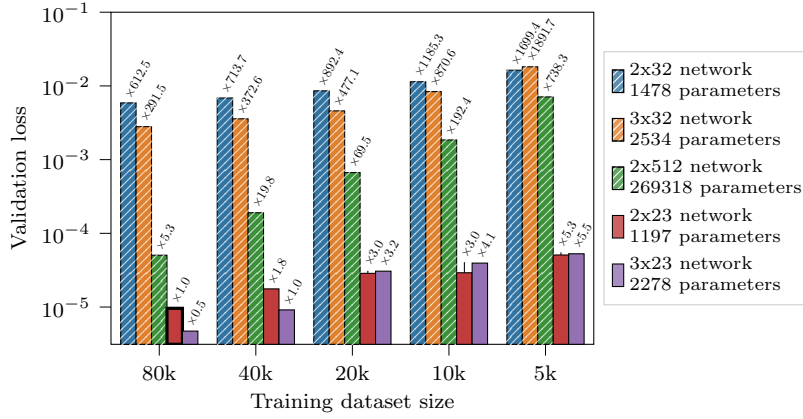

  \centering
  %
  \caption{%
    \emph{Final validation loss of neural networks trained on full and reduced neo-Hookean datasets.}
      The validation loss of TFENNs and standard networks is shown by solid and hatched bars, respectively.
      The ratio of the achieved validation loss to the one achieved by the tensor feature-based network with the least number of parameters, highlighted with a bar with a wider edge, is shown by the numbers above the bars.
  }%
  \label{fig:neohookean_results}
\end{figure}

A dataset of $\bm{C}$-$\bm{S}$ pairs was generated by sampling random deformation gradients $\bm{F}$ as described in \cref{sec:results:implementation:strain}, and by subsequently computing $\bm{C} = \bm{F}^\trp \bm{F}$, and $\bm{S}$ using \cref{eq:neo_hookean:constitutive_model}.
Different architectures of neural networks with both scalar features and tensor features (using isotropic symmetry-preserving parameters) were trained on a dataset of \num{20000} samples.
These networks are compared with respect to the validation loss measured on a distinct dataset of \num{4000} samples during training in \cref{fig:neohookean_training}.
It can be seen that TFENNs achieve a significantly lower validation loss than standard networks with a lower number of parameters.
These neural networks were also trained on a smaller training sets of \num{10000} and \num{5000} samples and on larger ones of \num{40000} and \num{80000} samples.
The validation loss obtained at the end of training for these sets is shown in \cref{fig:neohookean_results}.
An important observation is that the decrease in performance of TFENNs for fewer training samples is less significant than the one of standard networks.
Even with a dataset of \num{5000} samples, TFENNs achieve a validation loss comparable to the one obtained with the best standard neural network trained on \num{80000} samples.

To demonstrate the ability of TFENNs to preserve material symmetries to a much higher degree than standard networks, the trained networks were given random input tensor and random rotated versions of these inputs, and the resulting predictions were compared numerically.
Given a neural network model $f$, a set of $N$ random input tensors $\left\{\bm{C}_1, \dots, \bm{C}_N\right\}$, and a set of $N$ random rotation matrices $\left\{\bm{R}_1, \dots, \bm{R}_N\right\}$ from the material symmetry group, the resulting error in symmetry enforcement is given by
\begin{equation}\label{eq:neo_hookean:symmetry_error}
  \epsilon_{\mathrm{sym}} =
\frac{2}{N} \sum_{i=1}^N 
\frac{\left\| f\left(\bm{C}_i\right) - \bm{R}_i f\left(\bm{R}_i^\trp \bm{C}_i \bm{R}_i\right) \bm{R}_i^\trp \right\|_2}
{\left\| f\left(\bm{C}_i\right) \right\|_2 + \left\| f\left(\bm{R}_i^\trp \bm{C}_i \bm{R}_i \right)\right\|_2}.
\end{equation}
The results of this test with $N=200$ on different neural networks are shown in \cref{tab:neohookean_symmetry_test}.
It can be seen that traditional neural networks only marginally enforce material symmetries after training, with only a modest improvement of the symmetry error in comparison to the randomly initialized networks before training.
On the other hand, TFENNs achieve a symmetry error that is on par with the numerical precision of the computations, demonstrating that material symmetries are enforced to a high degree.

\begin{table}[h]
  \centering
  \caption{%
    Symmetry enforcement error of neural networks trained on the neo-Hookean dataset.
  }%
  \label{tab:neohookean_symmetry_test}
  \begin{tabular}{ll|rr}
    \multicolumn{2}{c|}{\textbf{Network}} & \multicolumn{2}{c}{\textbf{Symmetry error}} \\
    \textbf{Features} & \textbf{Architecture} & \textbf{Pre-training} & \textbf{Post-training} \\
    \hline
    \textbf{Scalar} & \num{3}x\num{32} & \num{9.26e-01} & \num{4.91e-01} \\
    \textbf{Scalar} & \num{2}x\num{32} & \num{1.29e+00} & \num{4.89e-01} \\
    \textbf{Scalar} & \num{3}x\num{64} & \num{1.37e+00} & \num{4.83e-01} \\
    \textbf{Scalar} & \num{2}x\num{64} & \num{1.59e+00} & \num{4.80e-01} \\
    \textbf{Scalar} & \num{3}x\num{128} & \num{9.08e-01} & \num{4.78e-01} \\
    \textbf{Scalar} & \num{2}x\num{128} & \num{1.34e+00} & \num{4.78e-01} \\
    \textbf{Tensor} & \num{3}x\num{23} & \num{2.39e-15} & \num{1.12e-15} \\
    \textbf{Tensor} & \num{2}x\num{23} & \num{1.26e-15} & \num{9.98e-16} \\
  \end{tabular}
\end{table}

\subsection{Tensegrity lattice unit cell}\label{sec:results:tensegrity}

\begin{figure}[h]
    \centering
    \begin{subfigure}[t]{0.45\textwidth}
      \centering
      \begin{tikzpicture}
        \tikzset{%
          pics/tcell/.style args={#1 #2 #3 #4}{%
            code = {%

            \tikzset{node/.style = {circle, fill, minimum size=3pt, inner sep=0pt, outer sep=0pt}}
            \tikzset{cable/.style = {blue,thick,line join=round,line cap=round,blend mode=darken}}
            \tikzset{bar/.style = {red,ultra thick,line join=round,line cap=round,blend mode=darken}}

              \def\l{4} 

              \begin{scope}[x={(#1,#2)}, y={(#3,#4)}]
                \draw[bar] ({-\l/3}, 0) -- ({\l/3}, 0);
                \draw[bar] (0, {-\l/3}) -- (0, {\l/3});

                \foreach\i in {1, -1}
                {%
                  \foreach\j in {1, -1}
                  {%
                    \draw[bar] ({\i*\l/2}, {\j*\l/6}) -- ({\i*\l/6}, {\j*\l/2});
                    \draw[bar] ({\i*\l/6}, {\j*\l/6}) -- ({\i*\l/2}, {\j*\l/2});
                  };
                };

                \foreach\i in {1, -1}
                {%
                  \foreach\j in {1, -1}
                  {%
                    \draw[cable] ({\i*\l/2}, {\j*\l/6}) -- ({\i*\l/6}, {\j*\l/6}) -- ({\i*\l/6}, {\j*\l/2})
                      -- ({\i*\l/2}, {\j*\l/2}) -- cycle -- ({\i*\l/3}, 0) -- ({\i*\l/6}, {\j*\l/6}) 
                      -- (0, {\j*\l/3}) -- ({\i*\l/6}, {\j*\l/2});
                  };
                };

                \foreach\i in {1, -1}
                {%
                  \foreach\j in {1, -1}
                  {%
                    \node[node] at ({\i*\l/6}, {\j*\l/6}) {};
                    \node[node] at ({\i*\l/6}, {\j*\l/2}) {};
                    \node[node] at ({\i*\l/2}, {\j*\l/6}) {};
                    \node[node] at ({\i*\l/2}, {\j*\l/2}) {};
                  };
                  \node[node] at ({\i*\l/3}, 0) {};
                  \node[node] at (0, {\i*\l/3}) {};
                };

              \end{scope}
            }
          }
        }
      \draw pic {tcell=0.7 0 0 0.7};
      \end{tikzpicture}
      \caption{Tensegrity cell~\cite{pirsoltan2018topics}. Bars and cables are
      respectively represented by thick red lines and thin blue lines.}%
      \label{fig:tensegrity_cell}
    \end{subfigure}
    \hfill
    \begin{subfigure}[t]{0.45\textwidth}
      \centering
      \includegraphics[width=0.53\textwidth]{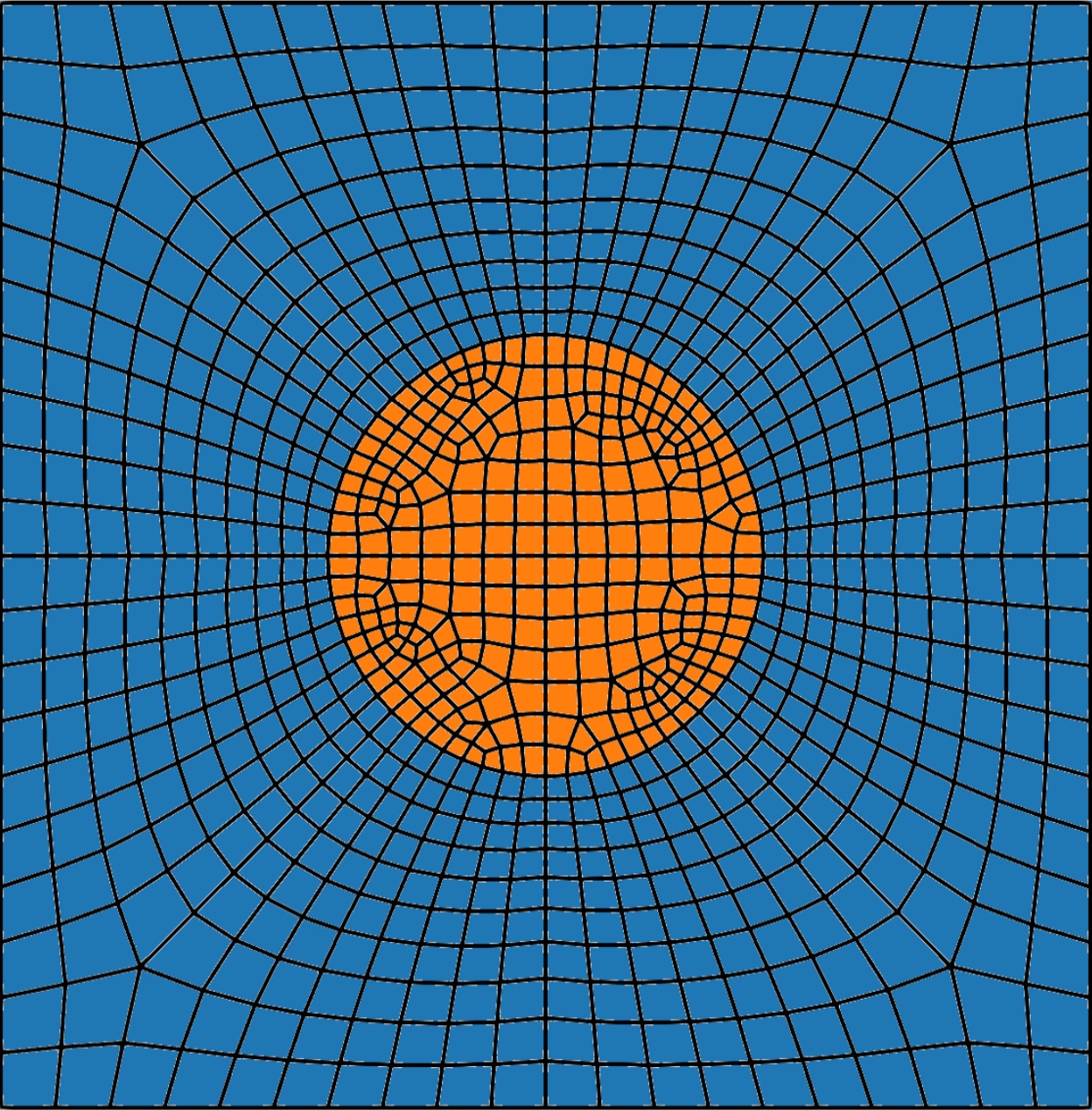}
      \caption{Mesh of the elasto-plastic microstructure used in finite elements analysis~\cite{logarzo2021smart}.}%
      \label{fig:elasto_plastic_microstructure}
    \end{subfigure}
    \caption{Unit cells used in the tensegrity lattice and elasto-plastic microstructure problems.}%
\end{figure}
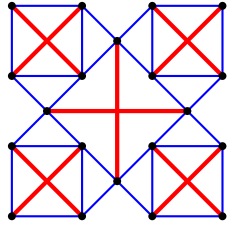
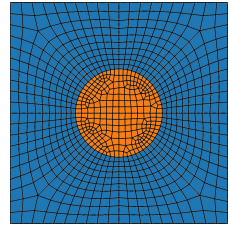

A two-dimensional unit cell of an elastic tensegrity metamaterial, shown in \cref{fig:tensegrity_cell}, is considered as a second example.
The cell is composed of pinned bars and cables.
Bars are modeled as linear elastic springs up to their Euler's buckling load, after which their load is assumed constant.
Cables are modeled as linear elastic springs which can only support tension.
For a given deformation gradient $\bm{F}$, the twelve nodes at the boundary of the cell are fixed according to $\bm{F}$ while the nodes inside the cell are unconstrained.
The equilibrium configuration of the cell is found by using a conjugate gradient method with a Polak-Ribière update formula~\cite{polak1969note}.
Given the boundary nodes' positions in the undeformed configuration $\bm{X} \in \mathbb{R}^{2\times 12}$, the forces at the nodes in the deformed configuration $\bm{f} \in \mathbb{R}^{2 \times 12}$, and the area of the undeformed cell $A_0$, the equivalent strain tensor is computed with
\begin{equation}\label{eq:cell_strain_tensor}
    \bm{S} = \frac{1}{A_0} \bm{X} \bm{f}^\trp \bm{F}^{-T}.
\end{equation}
\Cref{eq:cell_strain_tensor} is found by equating the formulations of the work done from deformations of the cell both in terms of strain/stress and displacements/forces.

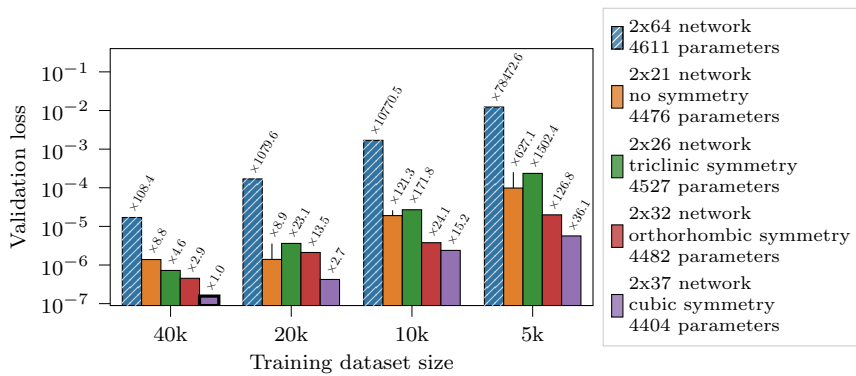
\begin{figure}[h]
  \centering
  \begin{tikzpicture}

  \definecolor{brown1926061}{RGB}{192,60,61}
  \definecolor{darkgray176}{RGB}{176,176,176}
  \definecolor{darkslategray66}{RGB}{66,66,66}

  \definecolor{lightgray204}{RGB}{204,204,204}
  \definecolor{mediumpurple147113178}{RGB}{147,113,178}
  \definecolor{peru22412844}{RGB}{224,128,44}
  \definecolor{seagreen5814558}{RGB}{58,145,58}
  \definecolor{steelblue49115161}{RGB}{49,115,161}

  \begin{axis}[
  height=5cm,
  legend cell align={left},
  legend style={
    fill opacity=0.8,
    draw opacity=1,
    text opacity=1,
    at={(1.02,0.5)},
    anchor=west,
    draw=lightgray204,
    font=\scriptsize,
    cells={align=left}
  },
  legend image code/.code={
    \draw [#1] (0cm, -0.15cm) rectangle (0.15cm, 0.15cm);
  },
  label style={font=\footnotesize},
  tick label style={font=\footnotesize},
  log basis y={10},
  tick align=outside,
  tick pos=left,
  unbounded coords=jump,
  width=8cm,
  x grid style={darkgray176},
  xlabel={Training dataset size},
  xmin=-0.5, xmax=3.5,
  xtick style={color=black},
  xtick={0,1,2,3},
  xticklabels={40k,20k,10k,5k},
  y grid style={darkgray176},
  ylabel={Validation loss},
  ymin=8.9251410800217e-08, ymax=0.4,
  ymode=log,
  ytick style={color=black},
  ytick={1e-09,1e-08,1e-07,1e-06,1e-05,0.0001,0.001,0.01,0.1,1},
  yticklabels={
    \(\displaystyle {10^{-9}}\),
    \(\displaystyle {10^{-8}}\),
    \(\displaystyle {10^{-7}}\),
    \(\displaystyle {10^{-6}}\),
    \(\displaystyle {10^{-5}}\),
    \(\displaystyle {10^{-4}}\),
    \(\displaystyle {10^{-3}}\),
    \(\displaystyle {10^{-2}}\),
    \(\displaystyle {10^{-1}}\),
    \(\displaystyle {10^{0}}\)
  }
  ]

  \draw[draw=black,fill=steelblue49115161,postaction={pattern=north east lines, pattern color=white, fill opacity=0.7}] (axis cs:-0.4,8.9251410800217e-08) rectangle (axis cs:-0.24,1.6998497997402e-05);
  \draw[draw=black,fill=steelblue49115161,postaction={pattern=north east lines, pattern color=white, fill opacity=0.7}] (axis cs:0.6,8.9251410800217e-08) rectangle (axis cs:0.76,0.000169289929586653);
  \draw[draw=black,fill=steelblue49115161,postaction={pattern=north east lines, pattern color=white, fill opacity=0.7}] (axis cs:1.6,8.9251410800217e-08) rectangle (axis cs:1.76,0.00168883204419262);
  \draw[draw=black,fill=steelblue49115161,postaction={pattern=north east lines, pattern color=white, fill opacity=0.7}] (axis cs:2.6,8.9251410800217e-08) rectangle (axis cs:2.76,0.012304640965105);

  \draw[draw=black,fill=peru22412844] (axis cs:-0.24,8.9251410800217e-08) rectangle (axis cs:-0.08,1.38426685418252e-06);
  \draw[draw=black,fill=peru22412844] (axis cs:0.76,8.9251410800217e-08) rectangle (axis cs:0.92,1.39581781670496e-06);
  \draw[draw=black,fill=peru22412844] (axis cs:1.76,8.9251410800217e-08) rectangle (axis cs:1.92,1.90141990997815e-05);
  \draw[draw=black,fill=peru22412844] (axis cs:2.76,8.9251410800217e-08) rectangle (axis cs:2.92,9.83308841983299e-05);

  \draw[draw=black,fill=seagreen5814558] (axis cs:-0.08,8.9251410800217e-08) rectangle (axis cs:0.08,7.23327803020714e-07);
  \draw[draw=black,fill=seagreen5814558] (axis cs:0.92,8.9251410800217e-08) rectangle (axis cs:1.08,3.62426582589703e-06);
  \draw[draw=black,fill=seagreen5814558] (axis cs:1.92,8.9251410800217e-08) rectangle (axis cs:2.08,2.69411651603102e-05);
  \draw[draw=black,fill=seagreen5814558] (axis cs:2.92,8.9251410800217e-08) rectangle (axis cs:3.08,0.00023557272554244);

  \draw[draw=black,fill=brown1926061] (axis cs:0.0799999999999999,8.9251410800217e-08) rectangle (axis cs:0.24,4.5296629077017e-07);
  \draw[draw=black,fill=brown1926061] (axis cs:1.08,8.9251410800217e-08) rectangle (axis cs:1.24,2.11114357416826e-06);
  \draw[draw=black,fill=brown1926061] (axis cs:2.08,8.9251410800217e-08) rectangle (axis cs:2.24,3.78330833553341e-06);
  \draw[draw=black,fill=brown1926061] (axis cs:3.08,8.9251410800217e-08) rectangle (axis cs:3.24,1.98881334707334e-05);

  \draw[draw=black,fill=mediumpurple147113178,very thick] (axis cs:0.24,8.9251410800217e-08) rectangle (axis cs:0.4,1.56801779716789e-07);
  \draw[draw=black,fill=mediumpurple147113178] (axis cs:1.24,8.9251410800217e-08) rectangle (axis cs:1.4,4.22346449185789e-07);
  \draw[draw=black,fill=mediumpurple147113178] (axis cs:2.24,8.9251410800217e-08) rectangle (axis cs:2.4,2.39034260077084e-06);
  \draw[draw=black,fill=mediumpurple147113178] (axis cs:3.24,8.9251410800217e-08) rectangle (axis cs:3.4,5.65841050161909e-06);

  \addlegendimage{ybar,draw=black,postaction={pattern=north east lines, pattern color=white, fill opacity=0.7},fill=steelblue49115161}
  \addlegendentry{2x64 network\\
  4611 parameters}

  \addlegendimage{ybar,draw=black,fill=peru22412844}
  \addlegendentry{2x21 network\\
  no symmetry\\
  4476 parameters
  }

  \addlegendimage{ybar,draw=black,fill=seagreen5814558}
  \addlegendentry{2x26 network\\
  triclinic symmetry\\
  4527 parameters}

  \addlegendimage{ybar,draw=black,fill=brown1926061}
  \addlegendentry{2x32 network\\
  orthorhombic symmetry\\
  4482 parameters}

  \addlegendimage{ybar,draw=black,fill=mediumpurple147113178}
  \addlegendentry{2x37 network\\
  cubic symmetry\\
  4404 parameters}

  \begin{scope}[
    every node/.append style={
      scale=0.5,
      anchor=south west,
      text=black,
      rotate=60,
      },
    ]
    \draw (axis cs:-0.32,1.6998497997402e-05) ++(2pt,-1pt) node{$\times108.4$};
    \draw (axis cs:0.68,0.000169289929586653) ++(2pt,-1pt) node{$\times1079.6$};
    \draw (axis cs:1.68,0.00168883204419262) ++(2pt,-1pt) node{$\times10770.5$};
    \draw (axis cs:2.68,0.012304640965105) ++(2pt,-1pt) node{$\times78472.6$};

    \draw (axis cs:-0.16,1.38426685418252e-06) ++(2pt,-1pt) node{$\times8.8$};
    \draw (axis cs:0.84,1.39581781670496e-06) -- ++(0,6pt) ++(2pt,-1pt) node{$\times8.9$};
    \draw (axis cs:1.84,1.90141990997815e-05) -- ++(0,2pt) ++(2pt,-1pt) node{$\times121.3$};
    \draw (axis cs:2.84,9.83308841983299e-05) -- ++(0,6pt) ++(2pt,-1pt) node{$\times627.1$};

    \draw (axis cs:0,7.23327803020714e-07) ++(2pt,-1pt) node{$\times4.6$};
    \draw (axis cs:1,3.62426582589703e-06) ++(2pt,-1pt) node{$\times23.1$};
    \draw (axis cs:2,2.69411651603102e-05) ++(2pt,-1pt) node{$\times171.8$};
    \draw (axis cs:3,0.00023557272554244) ++(2pt,-1pt) node{$\times1502.4$};

    \draw (axis cs:0.16,4.5296629077017e-07) ++(2pt,-1pt) node{$\times2.9$};
    \draw (axis cs:1.16,2.11114357416826e-06) ++(2pt,-1pt) node{$\times13.5$};
    \draw (axis cs:2.16,3.78330833553341e-06) ++(2pt,-1pt) node{$\times24.1$};
    \draw (axis cs:3.16,1.98881334707334e-05) ++(2pt,-1pt) node{$\times126.8$};

    \draw (axis cs:0.32,1.56801779716789e-07) ++(2pt,-1pt) node{$\times1.0$};
    \draw (axis cs:1.32,4.22346449185789e-07) ++(2pt,-1pt) node{$\times2.7$};
    \draw (axis cs:2.32,2.39034260077084e-06) ++(2pt,-1pt) node{$\times15.2$};
    \draw (axis cs:3.32,5.65841050161909e-06) ++(2pt,-1pt) node{$\times36.1$};
  \end{scope}
  \end{axis}

  \end{tikzpicture}
  \caption{%
    \emph{Final validation loss of neural networks trained on full and reduced tensegrity cell datasets with different enforced symmetries.}
    Plotting conventions as in \cref{fig:neohookean_results}.
  }%
  \label{fig:tcell_results}
\end{figure}

Here again, various neural networks were trained on different dataset sizes.
The resulting validation loss of neural networks trained on the tensegrity cell dataset is shown in \cref{fig:tcell_results}.
In addition, results with various enforced symmetries for TFENNs are provided.
TFENNs again outperform traditional neural networks with a similar number of parameters by a large margin.
Unsurprisingly, the best performing TFENN is the one with cubic symmetry enforced, as it is the most restrictive set of symmetries that the tensegrity cell verifies.
However, imposing orthotropic symmetry, which is less constraining but still verified by the tensegrity cell, still results in an improvement over a scalar feature formulation.
The same can be said when using a triclinic formulation, which does not enforce any symmetry but relies on weight tensors with major symmetry, and when using a completely unconstrained formulation of the tensor weights.
The difference in performance between the TFENNs with cubic symmetry and standard neural networks is even more pronounced for smaller datasets, showing again the high data efficiency of TFENNs.
It can be noted that in some cases, the TFENN with no enforced symmetry performs better than the one with triclinic or even orthotropic symmetry.
It is difficult to draw a general conclusion from this observation as this behavior is not consistent across all dataset sizes and the difference in performance is relatively small.
In any case, it is worth noting that these three formulations still largely outperform the standard network, therefore supporting the concept of neural networks based on tensor features.

A test was also performed with a TFENN enforcing isotropic symmetry, which is not verified by the tensegrity cell, and naturally resulted in the network being incapable of learning the cell's response.
This test is not represented in \cref{fig:tcell_results} as the final validation loss, ranging between \num{8.429e-01} and \num{1.13} for the various dataset sizes, is much higher than the displayed ones.
From this result, it can be guessed that in the case of unknown dataset symmetry, multiple TFENNs with different enforced symmetries may be trained, and the one with the lowest validation loss selected, thereby offering a mean of discovering a dataset symmetry.

\subsection{Elasto-plastic microstructure}\label{sec:results:elasto_plastic}

The last example considered is a two-dimensional microstructure composed of a hard elastic inclusion in a soft elastic-plastic matrix.
\num{10000} sequence samples of \num{200} loading steps each were generated via a finite element simulation, as detailed in~\cite{logarzo2021smart}.
The microstructure, which has cubic symmetry, is shown in \cref{fig:elasto_plastic_microstructure}.
In these cases, the input and output tensors are respectively the infinitesimal strain tensor and the average Cauchy stress tensor.

\begin{figure}[h]
  \centering
  \begin{tikzpicture}

  \definecolor{brown1926061}{RGB}{192,60,61}
  \definecolor{darkgray176}{RGB}{176,176,176}

  \definecolor{lightgray204}{RGB}{204,204,204}
  \definecolor{peru22412844}{RGB}{224,128,44}
  \definecolor{seagreen5814558}{RGB}{58,145,58}
  \definecolor{steelblue49115161}{RGB}{49,115,161}

  \begin{axis}[
  height=5cm,
  legend cell align={left},
  legend style={
    fill opacity=0.8,
    draw opacity=1,
    text opacity=1,
    at={(1.02,0.5)},
    anchor=west,
    draw=lightgray204,
    font=\scriptsize,
    cells={align=left}
  },
  legend image code/.code={
    \draw [#1] (0cm, -0.15cm) rectangle (0.15cm, 0.15cm);
  },
  label style={font=\footnotesize},
  tick label style={font=\footnotesize},
  log basis y={10},
  tick align=outside,
  tick pos=left,
  unbounded coords=jump,
  width=8cm,
  x grid style={darkgray176},
  xlabel={Training dataset size},
  xmin=-0.5, xmax=3.5,
  xtick style={color=black},
  xtick={0,1,2,3},
  xticklabels={8k,5k,3k,1k},
  y grid style={darkgray176},
  ylabel={Validation loss},
  ymin=0.000777500142152688, ymax=50,
  ymode=log,
  ytick style={color=black},
  ytick={1e-05,0.0001,0.001,0.01,0.1,1,10,100},
  yticklabels={
    \(\displaystyle {10^{-5}}\),
    \(\displaystyle {10^{-4}}\),
    \(\displaystyle {10^{-3}}\),
    \(\displaystyle {10^{-2}}\),
    \(\displaystyle {10^{-1}}\),
    \(\displaystyle {10^{0}}\),
    \(\displaystyle {10^{1}}\),
    \(\displaystyle {10^{2}}\)
  }
  ]

  \draw[draw=black,fill=steelblue49115161,postaction={pattern=north east lines, pattern color=white, fill opacity=0.7}] (axis cs:-0.4,0.000717463513879264) rectangle (axis cs:-0.2,0.0385527388104345);
  \draw[draw=black,fill=steelblue49115161,postaction={pattern=north east lines, pattern color=white, fill opacity=0.7}] (axis cs:0.6,0.000717463513879264) rectangle (axis cs:0.8,0.148658578329831);
  \draw[draw=black,fill=steelblue49115161,postaction={pattern=north east lines, pattern color=white, fill opacity=0.7}] (axis cs:1.6,0.000717463513879264) rectangle (axis cs:1.8,0.599060884450612);
  \draw[draw=black,fill=steelblue49115161,postaction={pattern=north east lines, pattern color=white, fill opacity=0.7}] (axis cs:2.6,0.000717463513879264) rectangle (axis cs:2.8,4.09147241991645);

  \draw[draw=black,fill=peru22412844,postaction={pattern=north east lines, pattern color=white, fill opacity=0.7}] (axis cs:-0.2,0.000717463513879264) rectangle (axis cs:2.77555756156289e-17,0.0257192647655184);
  \draw[draw=black,fill=peru22412844,postaction={pattern=north east lines, pattern color=white, fill opacity=0.7}] (axis cs:0.8,0.000717463513879264) rectangle (axis cs:1,0.090243197066037);
  \draw[draw=black,fill=peru22412844,postaction={pattern=north east lines, pattern color=white, fill opacity=0.7}] (axis cs:1.8,0.000717463513879264) rectangle (axis cs:2,0.454728762425717);
  \draw[draw=black,fill=peru22412844,postaction={pattern=north east lines, pattern color=white, fill opacity=0.7}] (axis cs:2.8,0.000717463513879264) rectangle (axis cs:3,3.3509967263644);

  \draw[draw=black,fill=seagreen5814558,postaction={pattern=north east lines, pattern color=white, fill opacity=0.7}] (axis cs:-2.77555756156289e-17,0.000717463513879264) rectangle (axis cs:0.2,0.013807710216014);
  \draw[draw=black,fill=seagreen5814558,postaction={pattern=north east lines, pattern color=white, fill opacity=0.7}] (axis cs:1,0.000717463513879264) rectangle (axis cs:1.2,0.0449496015085006);
  \draw[draw=black,fill=seagreen5814558,postaction={pattern=north east lines, pattern color=white, fill opacity=0.7}] (axis cs:2,0.000717463513879264) rectangle (axis cs:2.2,0.18190824311275);
  \draw[draw=black,fill=seagreen5814558,postaction={pattern=north east lines, pattern color=white, fill opacity=0.7}] (axis cs:3,0.000717463513879264) rectangle (axis cs:3.2,2.58673740903364);

  \draw[draw=black,fill=brown1926061,very thick] (axis cs:0.2,0.000717463513879264) rectangle (axis cs:0.4,0.0010830805664988);
  \draw[draw=black,fill=brown1926061] (axis cs:1.2,0.000717463513879264) rectangle (axis cs:1.4,0.00206429043589316);
  \draw[draw=black,fill=brown1926061] (axis cs:2.2,0.000717463513879264) rectangle (axis cs:2.4,0.00361926701998208);
  \draw[draw=black,fill=brown1926061] (axis cs:3.2,0.000717463513879264) rectangle (axis cs:3.4,0.0175216931734945);

  \addlegendimage{ybar,draw=black,postaction={pattern=north east lines, pattern color=white, fill opacity=0.7},fill=steelblue49115161}
  \addlegendentry{2x64 network\\
  38147 parameters}

  \addlegendimage{ybar,draw=black,postaction={pattern=north east lines, pattern color=white, fill opacity=0.7},fill=peru22412844}
  \addlegendentry{2x100 network\\
  92003 parameters}

  \addlegendimage{ybar,draw=black,postaction={pattern=north east lines, pattern color=white, fill opacity=0.7},fill=seagreen5814558}
  \addlegendentry{2x200 network\\
  364003 parameters}

  \addlegendimage{ybar,draw=black,fill=brown1926061}
  \addlegendentry{2x49 network\\
  36800 parameters}

  \begin{scope}[
    every node/.append style={
      scale=0.5,
      anchor=south west,
      text=black,
      rotate=60,
      },
    ]
    \draw (axis cs:-0.3,0.0385527388104345) ++(2pt,-1pt) node{$\times35.6$};
    \draw (axis cs:0.7,0.148658578329831) ++(2pt,-1pt) node{$\times137.3$};
    \draw (axis cs:1.7,0.599060884450612) ++(2pt,-1pt) node{$\times553.1$};
    \draw (axis cs:2.7,4.09147241991645) ++(2pt,-1pt) node{$\times3777.6$};

    \draw (axis cs:-0.1,0.0257192647655184) ++(2pt,-1pt) node{$\times23.7$};
    \draw (axis cs:0.9,0.090243197066037) ++(2pt,-1pt) node{$\times83.3$};
    \draw (axis cs:1.9,0.454728762425717) ++(2pt,-1pt) node{$\times419.8$};
    \draw (axis cs:2.9,3.3509967263644) ++(2pt,-1pt) node{$\times3093.9$};

    \draw (axis cs:0.1,0.013807710216014) ++(2pt,-1pt) node{$\times12.7$};
    \draw (axis cs:1.1,0.0449496015085006) ++(2pt,-1pt) node{$\times41.5$};
    \draw (axis cs:2.1,0.18190824311275) ++(2pt,-1pt) node{$\times168.0$};
    \draw (axis cs:3.1,2.58673740903364) ++(2pt,-1pt) node{$\times2388.3$};

    \draw (axis cs:0.3,0.0010830805664988) ++(2pt,-1pt) node{$\times1.0$};
    \draw (axis cs:1.3,0.00206429043589316) ++(2pt,-1pt) node{$\times1.9$};
    \draw (axis cs:2.3,0.00361926701998208) ++(2pt,-1pt) node{$\times3.3$};
    \draw (axis cs:3.3,0.0175216931734945) ++(2pt,-1pt) node{$\times16.2$};
  \end{scope}
  \end{axis}

  \end{tikzpicture}
    \caption{%
      \emph{Final validation loss of neural networks trained on full and reduced elasto-plastic microstructure datasets.}
      Plotting conventions as in \cref{fig:neohookean_results}.
    }%
    \label{fig:ep_results}
\end{figure}
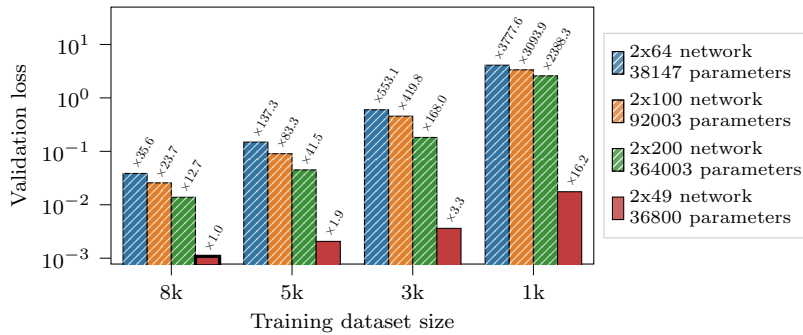

A TFENN enforcing cubic symmetry and multiple standard neural networks were trained on different sizes of datasets, up to \num{8000} samples, and were tested on the remaining \num{2000} samples.
All networks have two hidden layers and the TFENN has the least number of parameters.
The final validation loss of all networks on the different dataset sizes is reported in \cref{fig:ep_results}.
The conclusions from this example are similar to the ones obtained from previous experiments: TFENNs provide a significant improvement in data efficiency and prediction accuracy, with a reduced number of parameters.
The resulted observed so far therefore extend to history-dependent models.

\subsection{Discussion}

Three examples of various complexity were considered in this section, both for history-independent and history-dependent models.
The presented results show that TFENNs, which enforce material symmetries down to floating point precision, are capable of learning the response of various materials and microstructures with a high accuracy and a relatively low number of training samples.
TFENNs are particularly efficient in comparison to standard neural networks with small datasets.
Indeed, with all examples, the relative improvement of TFENNs over standard neural networks is more pronounced with fewer training examples.

Even when no symmetry is enforced and the tensor weights and biases are unconstrained, TFENNs still outperform standard neural networks, as shown with the case of the tensegrity cell.
Given that the only difference between TFENNs with no enforced symmetry and standard neural networks is the use of tensorial activation functions, this result suggests that these activation functions alone play a significant role in the performance of TFENNs.

\subsection{Symmetry basis discovery}\label{sec:results:symmetry_basis_discovery}

In all three previous examples, the basis of symmetry of the material was known and used in the implementation of the TFENNs.
However, in practice, the basis of symmetry of a material may not always be known, which may seem like a limitation of TFENNs.
To address this issue, a method to discover the basis of symmetry of a material from data is proposed in this section.
This method consists of transforming the input and output tensors of the TFENNs during training via a learnable transformation matrix.
That is, if the neural network operation is denoted by $f$, a $d \times d$ rotation matrix $\bm{R}$ is defined and the prediction of the network combined with the rotation is given by
\begin{equation}
  \hat{\bm{y}} = \bm{R} f(\bm{R}^\trp\bm{x}\bm{R}) \bm{R}^\trp.
\end{equation}
Clearly, if $f$ is constrained to enforce the material symmetries assuming according to a chosen arbitrary basis, for the resulting model to learn the material's response, then $\bm{R}$ has to be a rotation from the actual basis of symmetry of the material to the chosen basis for the representation of $f$.
To discover an appropriate matrix that satisfies this property, $\bm{R}$ is computed from a set of learnable parameters of the neural network and added during training.
In two dimensions, this parameter is a single angle $\theta$ and in three dimensions, a possible choice is a set of four parameters defining a quaternion $\bm{q}$.
For the three-dimensional case, $\bm{q}$ is normalized to obtain the resulting rotation matrix $\bm{R}$ and a term penalizing the deviation of $\bm{q}$ from a unit quaternion can be added to the loss function.

This method was tested on the tensegrity cell dataset, which was transformed into five different arbitrarily rotated versions of it.
A TFENN enforcing cubic symmetry with the added learnable rotation was trained five times with different random initial parameters on each of these datasets.
The final validation loss and the final value of the learned angle for each rotated dataset and initialization seed are reported in \cref{tab:rotation_results}.
It can be seen that all but one case result in a low final validation loss with a learned rotation close to the actual rotation of the dataset (up to a \SI{45}{\degree} rotation, which preserves the cubic symmetry of the tensegrity cell).
The model that fails to find the correct rotation also has a high final validation loss, making the failure easy to detect.
In addition, the initial angle difference, which depends on the random initialization seed, is reported in \cref{tab:rotation_results}, as it shows that the ability of the model to find the correct rotation is not particularly correlated with the magnitude of the initial angle difference.
Therefore, it can be concluded that this method is effective at discovering the basis of symmetry of a material from data with a few training experiments based on different random initializations.

\begin{table}[h]
  \centering
  \caption{%
    Final validation loss and learned rotation angle of TFENNs trained on rotated versions of the tensegrity cell dataset.
    The only failing case occurs for a rotation of \ang{20} and a random initialization seed of 2.
  }%
  \label{tab:rotation_results}
  \begin{tabular}{cc|ccc}
    \multirowcell{3}{\textbf{True}\\\textbf{rotation}} & \multirowcell{3}{\textbf{Seed}} & \multirowcell{3}{\textbf{Final}\\\textbf{validation}\\\textbf{loss}} & \multicolumn{2}{c}{\multirowcell{2}{\textbf{Predicted rotation (error)}}}  \\
                                                                       & & & & \\
                                               &                             &                                               & \textbf{Pre-training} & \textbf{Post-training} \\
    \hline
    \multirow{5}{*}{\ang{20}}                  & 0                           & \num{9.25e-06}                                & \ang{-31.727} (\ang{6.727}) & \ang{-24.993} (\ang{0.007}) \\
                                               & 1                           & \num{1.71e-04}                                & \ang{53.716} (\ang{11.284}) & \ang{64.964} (\ang{0.036}) \\
                                               & 2                           & \num{1.02e+00}                                & \ang{-139.904} (\ang{20.096}) & \ang{-142.341} (\ang{17.659}) \\
                                               & 3                           & \num{1.21e-05}                                & \ang{-16.344} (\ang{8.656}) & \ang{-25.005} (\ang{0.005}) \\
                                               & 4                           & \num{7.71e-05}                                & \ang{-41.065} (\ang{16.065}) & \ang{-24.983} (\ang{0.017}) \\
    \hline
    \multirow{5}{*}{\ang{175}}                 & 0                           & \num{1.57e-04}                                & \ang{-31.727} (\ang{18.273}) & \ang{-50.053} (\ang{0.053}) \\
                                               & 1                           & \num{4.68e-06}                                & \ang{53.716} (\ang{13.716}) & \ang{39.999} (\ang{0.001}) \\
                                               & 2                           & \num{1.38e-05}                                & \ang{-139.904} (\ang{0.096}) & \ang{-140.005} (\ang{0.005}) \\
                                               & 3                           & \num{7.99e-06}                                & \ang{-16.344} (\ang{11.344}) & \ang{-4.989} (\ang{0.011}) \\
                                               & 4                           & \num{1.12e-05}                                & \ang{-41.065} (\ang{8.935}) & \ang{-49.971} (\ang{0.029}) \\
    \hline
    \multirow{5}{*}{\ang{222}}                 & 0                           & \num{5.25e-05}                                & \ang{-31.727} (\ang{16.273}) & \ang{-47.999} (\ang{0.001}) \\
                                               & 1                           & \num{5.18e-05}                                & \ang{53.716} (\ang{11.716}) & \ang{42.009} (\ang{0.009}) \\
                                               & 2                           & \num{6.12e-05}                                & \ang{-139.904} (\ang{1.904}) & \ang{-137.993} (\ang{0.007}) \\
                                               & 3                           & \num{1.11e-04}                                & \ang{-16.344} (\ang{13.344}) & \ang{-2.994} (\ang{0.006}) \\
                                               & 4                           & \num{1.08e-04}                                & \ang{-41.065} (\ang{6.935}) & \ang{-47.999} (\ang{0.001}) \\
    \hline
    \multirow{5}{*}{\ang{298}}                 & 0                           & \num{4.83e-05}                                & \ang{-31.727} (\ang{14.727}) & \ang{-17.008} (\ang{0.008}) \\
                                               & 1                           & \num{2.43e-05}                                & \ang{53.716} (\ang{19.284}) & \ang{73.024} (\ang{0.024}) \\
                                               & 2                           & \num{1.76e-05}                                & \ang{-139.904} (\ang{12.096}) & \ang{-152.010} (\ang{0.010}) \\
                                               & 3                           & \num{3.76e-06}                                & \ang{-16.344} (\ang{0.656}) & \ang{-16.993} (\ang{0.007}) \\
                                               & 4                           & \num{1.12e-04}                                & \ang{-41.065} (\ang{20.935}) & \ang{-61.939} (\ang{0.061}) \\
    \hline
    \multirow{5}{*}{\ang{340}}                 & 0                           & \num{5.28e-06}                                & \ang{-31.727} (\ang{11.727}) & \ang{-20.000} (\ang{0.000}) \\
                                               & 1                           & \num{8.65e-06}                                & \ang{53.716} (\ang{16.284}) & \ang{70.001} (\ang{0.001}) \\
                                               & 2                           & \num{3.18e-05}                                & \ang{-139.904} (\ang{15.096}) & \ang{-155.011} (\ang{0.011}) \\
                                               & 3                           & \num{8.93e-06}                                & \ang{-16.344} (\ang{3.656}) & \ang{-19.999} (\ang{0.001}) \\
                                               & 4                           & \num{5.00e-03}                                & \ang{-41.065} (\ang{21.065}) & \ang{-19.971} (\ang{0.029}) \\
  \end{tabular}
\end{table}

\section{Conclusion}\label{sec:conclusion}

In this work, a novel type of neural network named TFENN was introduced.
These neural networks are designed to work with input-output pairs of symmetric tensors and to be equivariant to the action of subgroups of the orthogonal group, with the main application being the constitutive modeling of materials with symmetries.
This equivariance is exact for both finite and infinite groups, unlike some other methods that have been proposed in the literature.
TFENNs rely on a representation of the inputs, outputs, and intermediate features of the networks as symmetric tensors of same order and dimensionality.
The equivariance of the network is then enforced at the neuron-level by two techniques:
\begin{enumerate}
    \item Weights and biases are constrained to vector subspaces that depend on the desired enforced symmetry, a technique analogous to weight sharing which reduces the total number of free parameters.
    \item The activation functions are applied to the eigenvalues of their input tensors.
\end{enumerate}
These methods are easily extended to encompass the case of recurrent neural networks, which are used to model history-dependent problems.
Despite the fact that this method for enforcing equivariance may seem overly constraining, TFENNs are shown to be capable of learning complex constitutive models on a variety of material models and geometries, elastic or inelastic.
In fact, the various examples presented in this work show that TFENNs perform better than standard fully-connected neural networks, in particular when the training data is scarce.
This better performance can intuitively be partially attributed to the fact that TFENNs preserve material symmetries by construction and thus do no need to learn them from data, which means that the learning capacity of the network is fully utilized toward learning the constitutive model instead.
Since when no particular symmetry is enforced, TFENNs still tend to outperform traditional neural networks, the role of tensor-wise activation functions seem to be of particular importance.
A possible avenue for future research is thus to investigate the effect of these activation functions on new problems, particularly for materials with no symmetry, since they could not benefit from the restricted linear operations used on materials with symmetries.
The impact of imposing the major symmetry of the weight tensors is also still unclear, and could be investigated further, particularly for materials that are not hyperelastic.
Another promising direction would be to combine the methodology presented in this work with techniques that enforce other material properties, such as potential networks used to learn strain energy fields or input convex neural networks for enforcing material stability.
Finally, given the fairly general nature of the proposed method and its successful application to constitutive modeling, it would be interesting to apply it to other fields involving tensorial data with symmetry properties.

\section*{Acknowledgments}

Funding for this paper was provided by the U.S. Air Force Office of Scientific Research under grant number FA9550-24-1-0014.

\section*{Data availability}

The implementation of TFENNs' neurons is available at \url{https://github.com/kgaranger/TFENN}.
The implementation of the training procedures along with the data used in this work are available at \url{https://github.com/kgaranger/TFENN-examples}.

\printbibliography

\end{document}